\documentclass[letterpaper, 10pt, conference]{ieeeconf}  

\IEEEoverridecommandlockouts                              

\usepackage{cite}

\usepackage{graphics} 
\usepackage{epsfig} 
\usepackage{mathptmx} 
\usepackage{xcolor}
\usepackage{caption}
\usepackage{subcaption}
\usepackage{graphicx}
\usepackage[export]{adjustbox}
\usepackage{nameref}
\usepackage{hyperref}
\usepackage{amsmath}
\hypersetup{%
 setpagesize=false,%
 bookmarks=true,%
 bookmarksdepth=tocdepth,%
 bookmarksnumbered=true,%
 colorlinks=true,%
 linkcolor=blue,
 citecolor=blue,
 urlcolor=blue,
 pdftitle={},%
 pdfsubject={},%
 pdfauthor={},%
 pdfkeywords={}}

\usepackage[capitalise]{cleveref}

\title{\LARGE \bf
Learning Multimodal Attention for\\ Manipulating Deformable Objects with Changing States
}

\author{Namiko Saito$^{1,2}$, Mayu Tatsumi$^{3}$, Ayuna Kubo$^{3}$, Kanata Suzuki$^{1,4}$, Hiroshi Ito$^{1,5}$,\\ Shigeki Sugano$^{6}$ and Tetsuya Ogata$^{6,7}$ 
\thanks{*This work was supported by JST Moonshot R and D, Grant No. JPMJMS2031.}
\thanks{$^{1}$Authors are with Future Robotics Organization, Waseda University, Tokyo, Japan. $^{2}$Author is Microsoft Research Asia, Tokyo, Japan. ({\tt\small namikosaito@microsoft.com})
$^{3}$Authors are with Department of Modern Mechanical Engineering, Waseda University, Tokyo, Japan.
$^{4}$Author is with the Artificial Intelligence Laboratories, Fujitsu Limited, Kanagawa, Japan.
$^{5}$Author is with the Center for Technology Innovation - Controls and Robotics, Research \& Development Group, Hitachi, Ltd., Ibaraki, Japan.
$^{6}$Authors are with Faculty of Science and Engineering, Waseda University, Tokyo, Japan.
$^{7}$Author is with the National Institute of Advanced Science and Technology, Tokyo, Japan.}}

\begin{document}

\maketitle
\thispagestyle{empty}
\pagestyle{empty}

\begin{abstract}
To support humans in their daily lives, robots are required to adapt to objects whose states change due to external factors such as heat and force, and perform appropriate actions accordingly. 
Many objects in everyday environments exhibit such dynamic and continuous changes in their physical properties.
In these situations, sensory input from multiple modalities often contains both valuable and noisy information, and the importance of each sensor modality can shift over time as the object's state changes. 
This makes real-time perception and motion generation particularly challenging.
We propose a predictive recurrent neural network with an attention mechanism that dynamically weights sensor modalities based on their current reliability and relevance, enabling robots to achieve efficient perception and adaptive manipulation of objects undergoing state changes.
To demonstrate the effectiveness of the proposed method, we validated it on a physical humanoid robot, using a manipulation task of cooking scrambled eggs as an example scenario.
Our code and dataset are available here: \url{https://github.com/namikosaito/CookingScrambledEgg}

\end{abstract}


\section{Introduction}

\begin{figure}[]
\centering
\includegraphics[keepaspectratio, scale=0.5]{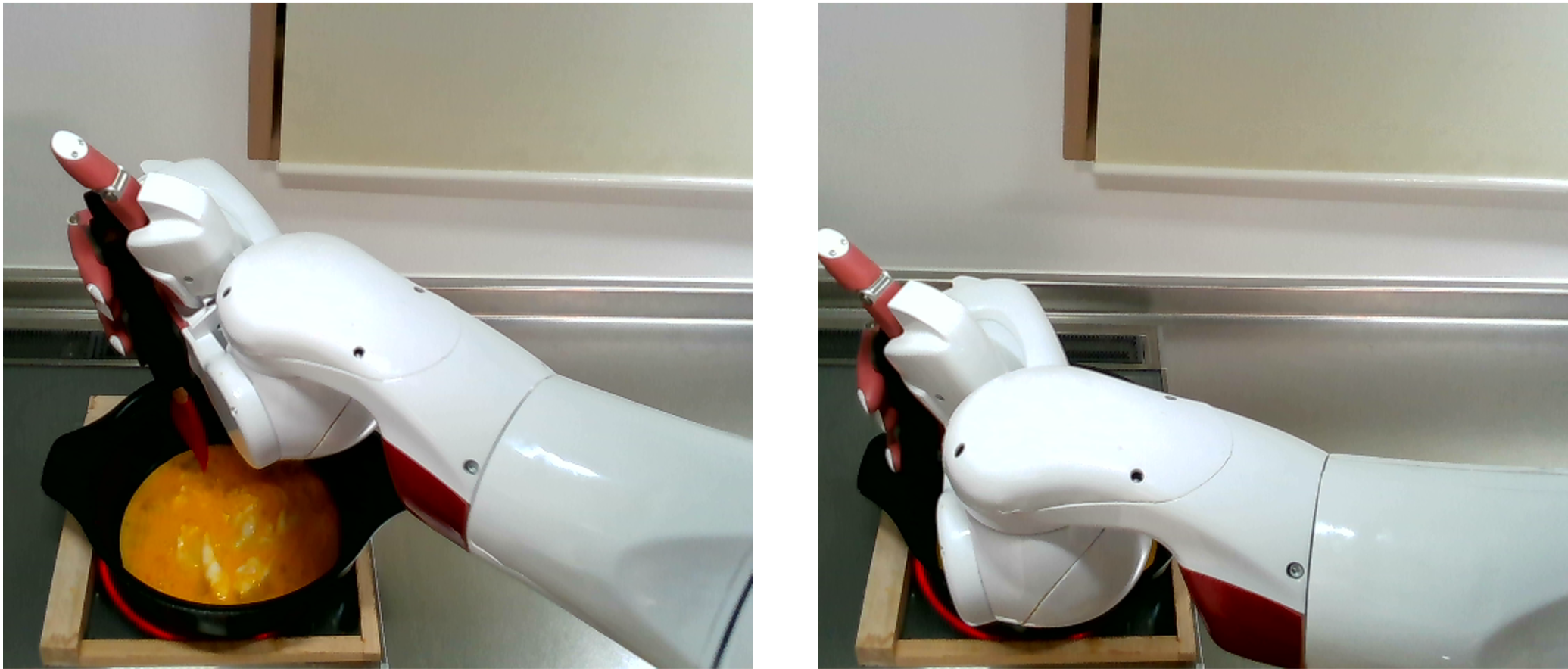}
\caption{
When the robot does not touch the pot (left), visual information is important. When the view is occluded by the arm (right), tactile and torque information is more reliable.
}
\label{f:approach}
\end{figure}

There is an increasing demand for robots that can support a wide variety of daily tasks.
To function in dynamic environments, robots must perceive changes in object states and adapt their actions in real time.
In many daily scenarios, objects are not static — they are affected by external factors such as heat and force, which cause both extrinsic properties (shape, color, position, etc.) and intrinsic properties (stiffness, friction, weight, etc.) to change over time~\cite{Susan1987, Dahiya2010, Bhattacharjee2019}.
Manipulating such objects presents unique challenges because materials may soften, harden, deform, or change phase over time, making it difficult to rely on predetermined motion plans.
Recognizing these changes requires the integration of multimodal sensory information, vision, tactile, and force.
However, this information often contains a mixture of valuable and noisy signals, and the importance of each sensor modality can fluctuate depending on the current state of the object.
For example, when the robot does not touch the pot, visual
information is important, while when the view is occluded, tactile and torque information become more reliable, as shown in Fig.~\ref{f:approach}.

Previous studies have mainly focused only on extrinsic properties~\cite{Yang2010, Petit2017, Lin2022, Salekin2019}.
Some studies showed intrinsic property recognition to be important,
however, they required pre-defined exploratory motions to recognize intrinsic ones before executing the main task~\cite{Saito2021_ral, Saito2019, LopezGuevara2020}, or relied on manually annotated labels~\cite{Gao2023}.
Other works using multimodal sensing handled stable or slowly-changing targets~\cite{Gemici2014, Sundaresan2022}.
We hypothesize that the key limitation is the inability to process sensory information quickly and efficiently, and generate adaptive motions in real time for rapidly changing objects.
To address this, we propose a predictive deep learning model with an attention mechanism that dynamically weights sensory modalities based on their current reliability and relevance.
While attention mechanisms have been applied to highlight important regions in images~\cite{Fukui2019, Ichiwara2022}, we extend this concept to multimodal sensor weighting during continuous manipulation tasks.
Our approach actively determines which information to focus on, and which to discard, guided by predictions of object behavior.

\begin{figure}[t]
\centering
\includegraphics[keepaspectratio, scale=0.6]{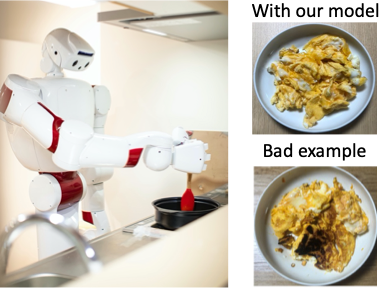}
\caption{
Cooking scrambled eggs.
}
\setlength\intextsep{0pt}
\label{f:AIREC}
\vspace*{-0.2cm}
\end{figure}
We demonstrate this approach in a scenario of scrambled egg cooking, where both vision and tactile/force feedback are essential to perceive dynamic object states and the robot must adjust its stirring method and direction in response to the changing properties of the egg.
The properties of egg mixture constantly change due to heating and stirring forces — becoming thicker, harder, clumpier, and more fragile over time.
The robot must perceive this evolving state and adjust its stirring speed, trajectory, and force accordingly, otherwise resulting in a poorly made dish, as shown in Fig.~\ref{f:AIREC}.
Relying solely on vision risks missing changes in texture or overcooking due to occlusion by the robot’s own arm, while tactile or force feedback alone cannot capture visual cues of burning.
Thus, scrambled eggs pose a challenging real-world case for handling rapidly changing object states.

\section{Method}
\subsection{Learning from Demonstration}
To enable robots to perform dexterous daily manipulation, we employ Learning from Demonstration (LfD)\cite{Ahmed2017, Ravichandar2020}, a common strategy in dexterous manipulation tasks\cite{Liu2022, Kim2022, Saigusa2022, Zhao2023, Fu2024}.
We collect expert demonstrations through teleoperation and train the robot using these data.

In our scenario, the robot is required to cook scrambled eggs while satisfying the following conditions: (1) Separate egg blocks so that no block exceeds 15 cm in length, (2) avoid burning, judged based on color (3) ensure no raw part remains.
Meeting these conditions simultaneously is highly challenging.
Without appropriate stirring in depth, only the bottom cooks, leaving raw areas on top.
If the egg is not mixed widely in the pot, raw parts reattach to become bigger blocks and burn unevenly.
Thus, the robot must adjust its motions while considering the distribution, softness, and thickness of egg blocks in real-time.
When simply replaying demonstration trajectories or moving randomly, the robot failed to achieve the goals — leaving raw areas, burning parts, or failing to separate large blocks.
Even replaying the original teleoperated trajectories did not succeed due to dynamic changes in egg state during cooking.
This highlights the necessity of real-time perception and motion adaptation rather than static, pre-recorded actions.

\subsection{Proposed learning model}
\begin{figure*}[t]
\centering
\includegraphics[keepaspectratio, scale=0.3]{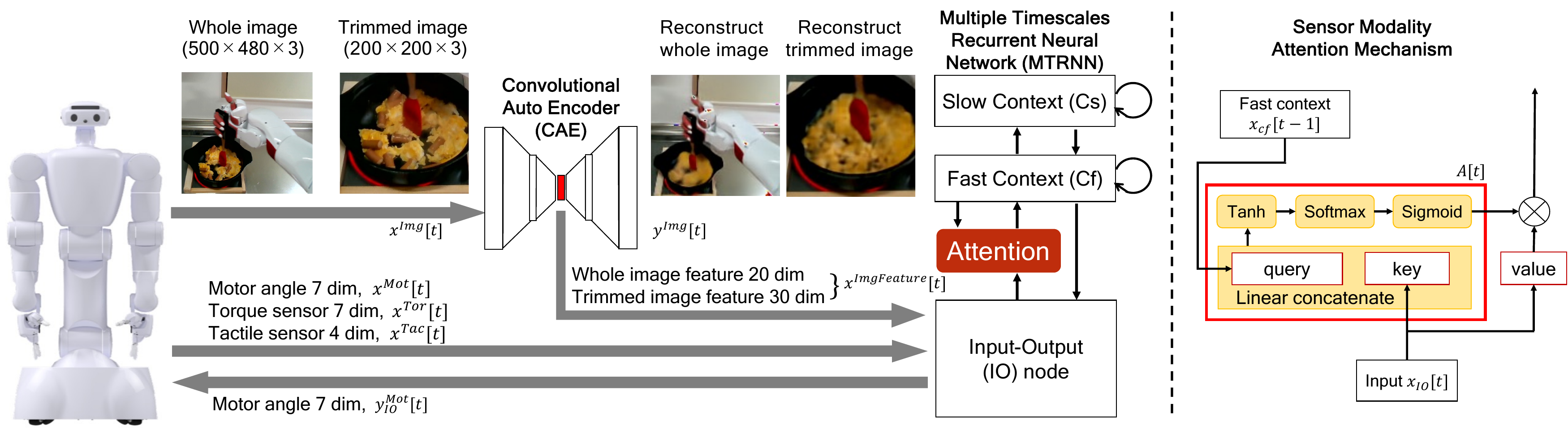}
\caption{
The proposed learning model (left) is composed of a CAE extracting image features, and a MTRNN with the sensor modality attention mechanism (right) conducting predictive learning considering the importance of each sensor modality.
}
\label{f:model}
\end{figure*}
\begin{table}[]
	\begin{center}
		\caption{Structure of the CAE}
 \begin{tabular}{lcccc} \hline \hline
 \textbf{} & \textbf{Input}& \textbf{Output} & \textbf{Activation} & \textbf{Processing} \\
    \hline
        1&(128, 128, 3)&(64, 64, 8)&ReLU&convolution\\
        2&(64, 64, 8)&(32, 32, 16)&ReLU&convolution\\
        3&(32, 32, 16)&(16, 16, 32)&ReLU&convolution\\
        4&8,192&1,000&ReLU&fully connected\\
        5&1,000&MID&Sigmoid&fully connected\\
        6&MID&1,000&ReLU&fully connected\\
        7&1,000&8,192&ReLU&fully connected\\
        8&(16, 16, 32)&(32, 32, 16)&ReLU&deconvolution\\
        9&(32, 32, 16)&(64, 64, 8)&ReLU&deconvolution\\
        10&(64, 64, 8)&(128, 128, 3)&ReLU&deconvolution\\ \hline
        \end{tabular}\\
        MID = {20: whole images, 30: trimmed images}
    \label{t:CAE_model}
    \vspace*{-0.2cm}
	\end{center}
\end{table}

\begin{table}[]
	\begin{center}
		\caption{Structure of the MTRNN}
 \begin{tabular}{ccc} \hline \hline
            Nodes & Time constant & Number of nodes \\ \hline
			${\rm {C_s}}$ & 32 & 7 \\ 
			${\rm {C_f}}$ & 5 & 30 \\ \hline
        \end{tabular}
    \label{t:MTRNN_model}
	\end{center}
\end{table}
Fig.\ref{f:model} shows the proposed deep learning model, composed of a convolutional autoencoder (CAE)\cite{Masci2011} for image feature extraction and a multiple timescales recurrent neural network (MTRNN)\cite{Yamashita2008} for motion generation.
The combination of CAE and MTRNN has been shown effective for motion generation in previous work\cite{Saito2021_ral}.
We adopted MTRNN for two main reasons: First, it provides multiple hierarchical context nodes with different timescales, which allows interpretable representation of long-term and short-term dependencies in task execution.
Second, its hierarchical structure is well-suited for filtering multimodal sensor inputs, where each modality (e.g., vision, touch, torque) can be weighted by its task relevance.
To enable this, we integrated a modality attention mechanism into the MTRNN, which dynamically adjusts the weights of sensorimotor inputs and passes only relevant information to each context node.

\subsubsection{CAE}
As for input, we use two different sizes of images, one shows the whole arm movement and the other is trimmed one which shows inside the pan.
We use the CAE to compress images to low-dimensional features such that the MTRNN learned all sensorimotor data in a well-balanced manner.
The CAE is trained to output reconstructed data (\(y^{\rm {Img}}[t]\)) identical to the input data (\(x^{\rm {Img}}[t]\)), where \(t\) is time step, by minimizing the mean squared error (MSE) shown below with the optimizer for the Adaptive moment estimation (Adam) algorithm. 
\begin{equation}
E=\sum_t\left(y^{\rm {Img}}[t]-x^{\rm {Img}}[t]\right)^2
\label{y_cae}
\end{equation}
We use the value of the intermediate layer as the image features (\(x^{\rm {ImgFeature}}[t]\)).

The Table.~\ref{t:CAE_model} shows the structural parameters used for constructing the CAE.
We set the number of the intermediate layer neurons to 20 for the whole images and to 30 for the trimmed images.
We tested changing the numbers of the intermediate layer nodes by 15, 20,..., 35 and set the number so that it was the smallest number in which output (\(y^{\rm {Img}}[t]\)) could reconstruct the original image (\(x^{\rm {Img}}[t]\)) by inspection.
The reconstructed images are not used for control but for experimenters to verify if the CAE represented the robot arm position by the whole images and the individual egg blocks by the trimmed images.
This module with CAE was trained for 1,500 epochs. 

\subsubsection{MTRNN}
We use the MTRNN as the main module to integrate all sensorimotor data, recognize the current egg states, and generate motions accordingly.
The MTRNN is a recurrent neural network that predicts the next step from the current and previous inputs. 
It has multiple nodes with different time constants as shown in Table.~\ref{t:MTRNN_model}. 
The Fast Context (${\rm {C_f}}$) nodes learn short-term primitive data with their small time constant, while the Slow Context (${\rm {C_s}}$) nodes learn sequential information and behave as the latent space with a large time constant.
We tested and selected time constants and number of nodes, which are the best combinations to minimize the error which will be described in Eq.~\ref{y}.

In this research, we implement a sensor modality attention mechanism in the MTRNN.
We make the value of IO nodes (\(x_{\rm {IO}} [t]\)) as the ``key'' and the ``value,'' and the value of ${\rm {C_f}}$ nodes in the previous step (\(x_{\rm {C_f}}[t-1]\)) as the ``query'' respectively.
Using the key and the query, the attention mechanism (\(A[t]\)) is calculated and learned as follows and represents the map of the focusing factor of the key.
\begin{equation}
u_A[t] = w_{A,{\rm{IO}}}[x_{\rm{IO}}
[t], x_{\rm {C_f}}[t-1]]
\end{equation}
\begin{equation}
y_A[t] = {\rm {tanh}}\left(u_A[t]\right)
\end{equation}
\begin{equation}
\label{attention_map_eq}
A[t] = {\rm {sigmoid}}\big({\rm {softmax}}(y_A[t])\big),
\end{equation}
where \(w_{A,{\rm{IO}}}\) is the learning weight, which is multiplied with concatenated key and query.
In Eq. (4), using only softmax makes most value ends up to be nearly zero, which makes only stored information from a few specific modalities in the context nodes.
We apply sigmoid function on the top of softmax to smooth the attention map and monitor all the modalities.

Forward calculations are described as follows. 
First, input data \({\rm {T}}[t]\), which is either the training data during model training or the real-time data during evaluation experiments.
The input data is the image features, which is output from CAE (\(x^{\rm {ImgFeature}}[t]\)), torque sensor data (\(x^{\rm {Tor}}[t]\)), tactile sensor data (\(x^{\rm {Tac}}[t]\)), and motor angle data (\(x^{\rm {Mot}}[t]\)). 
Input to IO nodes of MTRNN (\(x_{\rm{IO}}[t]\)) is combination of \({\rm {T}}[t]\) and prediction from the previous step, which is described later in Eq. (10).
\begin{equation}
T[t] = [x^{\rm {ImgFeature}}[t], x^{\rm {Tor}}[t], x^{\rm {Tac}}[t], x^{\rm {Mot}}[t]]
\end{equation}
The internal value \(u_i\) of the neuron \(i\) \((\in{\rm {IO, Cf, Cs}})\) at step \(t\) is calculated as follows.
For calculating internal value of ${\rm {C_f}}$ nodes (\(u_{\rm {C_f}}[t]\)), we utilize the attention mechanism multiplied with the ``value'' (\(x_{\rm{IO}}[t]\)).
\begin{equation}
u_{\rm {IO}}[t] = w_{{\rm {IO, Cf}}} x_{\rm {Cf}}[t]
\end{equation}
\begin{equation}
\begin{split}
u_{\rm {Cf}}[t] = 
& \big(1-\frac{1}{\tau_{\rm {Cf}}}\big)u_{\rm {Cf}}[t-1] + \frac{1}{\tau_{\rm {Cf}}}\big(w_{{\rm {Cf, IO}}} (A[t]x_{\rm {IO}}[t]) \\
& + w_{{\rm {Cf, Cs}}} x_{\rm {Cs}}[t] + w_{{\rm {Cf, Cf}}} x_{\rm {Cf}}[t]\big)
\end{split}
\end{equation}
\begin{equation}
u_{\rm {Cs}}[t] = \big(1-\frac{1}{\tau_{\rm {Cs}}}\big)u_{\rm {Cs}}[t-1] + \frac{1}{\tau_{\rm {Cs}}}\big(w_{{\rm {Cs, Cf}}} x_{\rm {Cf}}[t] + w_{{\rm {Cs, Cs}}} x_{\rm {Cs}}[t]\big)
\end{equation}
where \(\tau_i\) is the time constant of node \(i\), \(w_{i,j}\) is the weight value from node \(j\) to node \(i\), and \(x_j[t]\) is the input value. 
Subsequently, the output value is calculated as
\begin{equation}
y_i[t] = {\rm {tanh}}\left(u_i[t]\right).
\end{equation}
The module can generate robot motions by controlling the movements of the robot according to the output motor angle (\(y^{\rm {Mot}}_{\rm {IO}}[t]\)). 
Subsequently, the value of \(y_i[t]\) is used as the next input value, which is expressed as
\begin{equation}
x_i[t+1]= \left\{ 
\begin{array}{ll}
\alpha\times y_i[t]+(1-\alpha)\times {\rm {T}}[t+1] & i\in{{\rm {IO}}} \\\\
y_i[t] & i\in{{\rm {Cf, Cs}}}.
\end{array}\right.
\label{eq:x_i}
\end{equation}
The next input value \(x_{\rm {IO}}[t]\) is adjusted by multiplying the output of the preceding step \(y_{\rm {IO}}[t-1]\) and the data \({\rm {T}}[t]\) by the feedback rate \(\alpha\) (\(0 \le \alpha \le 1\)), which can adjust how much weight to give to its own predictions versus actual input. 
If the feedback rate is high, the model can stably predict the next step using own prediction history, whereas if the value is small, the model can flexibly adapt to real-time situations more sensitive to the real data.
We set 0.8 as the motor angle (\(x^{\rm {Mot}}_{\rm {IO}}[t+1]\)) and 0.6 for others (\(x^{\rm {ImgFeature}}_{\rm {IO}}[t+1]\), \(x^{\rm {Tor}}_{\rm {IO}}[t+1]\), \(x^{\rm {Tac}}_{\rm {IO}}[t+1]\)) which we empirically decided.

To realize backward calculations during training, the back propagation through time (BPTT) algorithm was used to minimize the training error as,
\begin{equation}
E=\sum_t\left(y_{{\rm {IO}}}[t-1]-{\rm {T}}[t]\right)^2.
\label{y}
\end{equation}
The weight is updated as
\begin{equation}
w^{n+1}_{ij}=w^{n}_{ij}-\eta \frac{\partial E}{\partial w^{n}_{ij}},
\label{weight}
\end{equation}
where \(\eta(=0.001)\) is the learning rate and \(n(=20,000)\) is the number of epochs with which the error was fully converged.

\subsection{Hardware and control}
We use Dry-AIREC, with 7-DOF dual arms and joint torque sensors.
An RGB camera (RealSense SR300) is mounted on its head, and FSR406 touch sensors are attached to the fingers and palm of its right hand.
The robot is impedance-controlled.

In a kitchen setting, the robot holds a pitcher with egg mixture in its left hand and a turner in its right.
The robot pours the egg mixture via a pre-defined motion, then stirs with the right hand controlled by the proposed learning model.

\subsection{Task setting and training data collection}

\begin{figure}[t]
\centering
\includegraphics[keepaspectratio, scale=0.4]{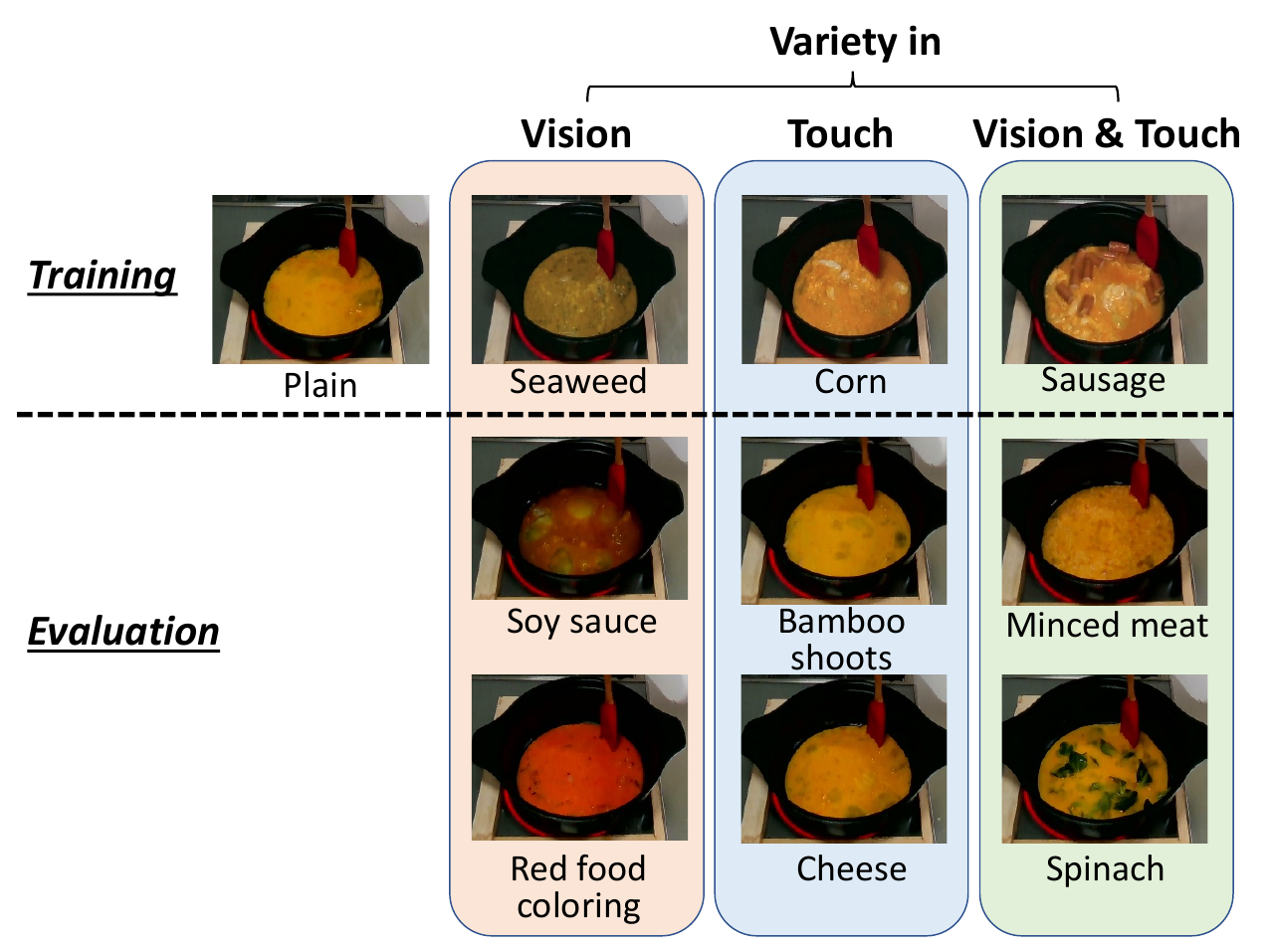}
\caption{
Variation in the egg mixture.
}
\label{f:egg}
\vspace*{-0.2cm}
\end{figure}

We conduct the experiment by changing the heating power, 180 and 190${}^\circ$C and the ingredients mixed in the egg mixture (Fig.~\ref{f:egg}).
Four mixtures are used for training: plain, with 1 g seaweed, 100 g corn, and 10 sausages — selected to isolate effects of vision, touch, or both.
For evaluation, we prepare six mixtures: ingredients affecting vision (soy sauce, food coloring), touch (bamboo shoots, cheese), and both (minced meat, spinach).

We collect demonstration dataset with teleoperation by commanding the end-effector position and solving inverse kinematics for deciding the arm configuration.
We took 32 dataset: (4 mixtures) × (2 temperatures) × (4 trials).
We sample at 2.5 Hz (every 0.4 s) and the average cooking time is 620 steps (248 s) at 180°C and 455 steps (182 s) at 190°C.
The right hand’s initial/final pose is fixed.
Input data to the learning model includes:
\begin{itemize}
    \item Images: whole (500 × 480 × 3), and trimmed (cropped 200 × 200 × 3 from its top left corner at (60, 270 pix)), resized to 128 × 128 × 3.
    \item Tactile sensors: 4 points (the index, middle and ring fingers and the palm)
    \item Torque sensors: 7 joints
    \item Motor angles: 7 joints
\end{itemize}
Image, tactile, torque, and motor angle data are normalized to [0, 1], [-0.85, 0.85], [-0.9, 0.9], and [-0.9, 0.9], based on vibration and noise levels.


\section{Result}
\subsection{Success rate of cooking}
\begin{table}[]
	\begin{center}
		\caption{Success rate of cooking with training ingredients}
		\begin{tabular}{c|c|c|c|c|c}\hline \hline
			heating power & plain & seaweed & corn & sausage & total\\ \hline
			180${}^\circ$C & 5/5 & 4/5 & 4/5 & 4/5 & 17/20\\ 
            &&(push out) & (block) & (block)&\\ \hline
			190${}^\circ$C & 5/5 & 4/5 & 5/5 & 5/5 & 19/20 \\ 
            &&(push out) &  & &\\ \hline
			total & 10/10 & 8/10 & 9/10 & 9/10 & 36/40\\ \hline
		\end{tabular}	\\
            ( ): reason of failure
		\label{t:success_rate_training}
        \vspace*{-0.2cm}
	\end{center}
\end{table}

\begin{table*}[]
	\begin{center}
		\caption{Success rate of cooking scrambled egg with UNTRAINED ingredients }
		\begin{tabular}{c|c|c|c|c|c|c|c}\hline \hline
            heating & \multicolumn{2}{c|}{vision} & \multicolumn{2}{c|}{touch} & \multicolumn{2}{c|}{vision \& touch} & total\\ \cline{2-7}
			power & say source & red food coloring & bamboo shoots & cheese & minced meat & spinach & \\ \hline
			180${}^\circ$C & 5/5 & 4/5 & 3/5 & 4/5 & 3/5 & 4/5 & 23/30\\ 
            &&(block) & (blocks) & (block)& (blocks) & (block)\\ \hline
			190${}^\circ$C & 4/5 & 5/5 & 4/5 & 2/5 & 5/5 & 4/5 & 24/30 \\ 
            &(block)&&(burn) & (push out, block, burn) & & (block)&\\ \hline
			total & 9/10 & 9/10 & 7/10 & 6/10 & 8/10 & 8/10 & 47/60\\ \cline{2-7}
            & \multicolumn{2}{c|}{18/20} &\multicolumn{2}{c|}{13/20}&\multicolumn{2}{c|}{16/20}& \\ \hline
		\end{tabular}	
            \\( ): reason of failure
		\label{t:success_rate_test}
        \vspace*{-0.2cm}
	\end{center}
\end{table*}
Table.~\ref{t:success_rate_training} shows the success rate of cooking with the trained egg mixture. 
We performed 40 trials, and in total, 36 (90.0\%) trials met all rules: (1) no egg block larger than 15 cm, (2) no burned parts, and (3) no raw parts. 
The table also shows the reasons for failure.
With the plain egg mixture, which is the simplest one, the robot succeeded in all trials.
While with seaweed, the robot pushed out the pan in two trials, which is because the color of seaweed is dark blue, the CAE had difficulty distinguishing between the seaweed and the black pot.
In the trial with a heating power of 180${}^\circ$C, one trial each with corn and sausage, some large blocks remained in the pan and failed.
Cooking at a heating power of 180${}^\circ$C is more difficult than cooking at 190${}^\circ$C. 
At 180${}^\circ$C, it takes more time for the egg to harden and the egg can easily stick to the other egg blocks again even after the robot has split the egg block.

Table.~\ref{t:success_rate_test} shows the results of the robustness evaluation using untrained ingredients.
We performed 60 trials and, in total, succeeded 47 (78.3\%) times.
The way to stir and the time to cook differ depending on the ingredients; however, the robot with our learning model recognized the properties and flexibly adjusted them.
The egg with cheese trial was especially difficult because the cheese is sticky and stretchy and its properties are considerably different from those of the ingredients used for training; however, the model succeeded in 6/10 trials.
We concluded that the model has acquired good generalization ability.

\subsection{Action change depends on egg status}
\begin{figure*}[h]
\centering
\includegraphics[width=5.in]{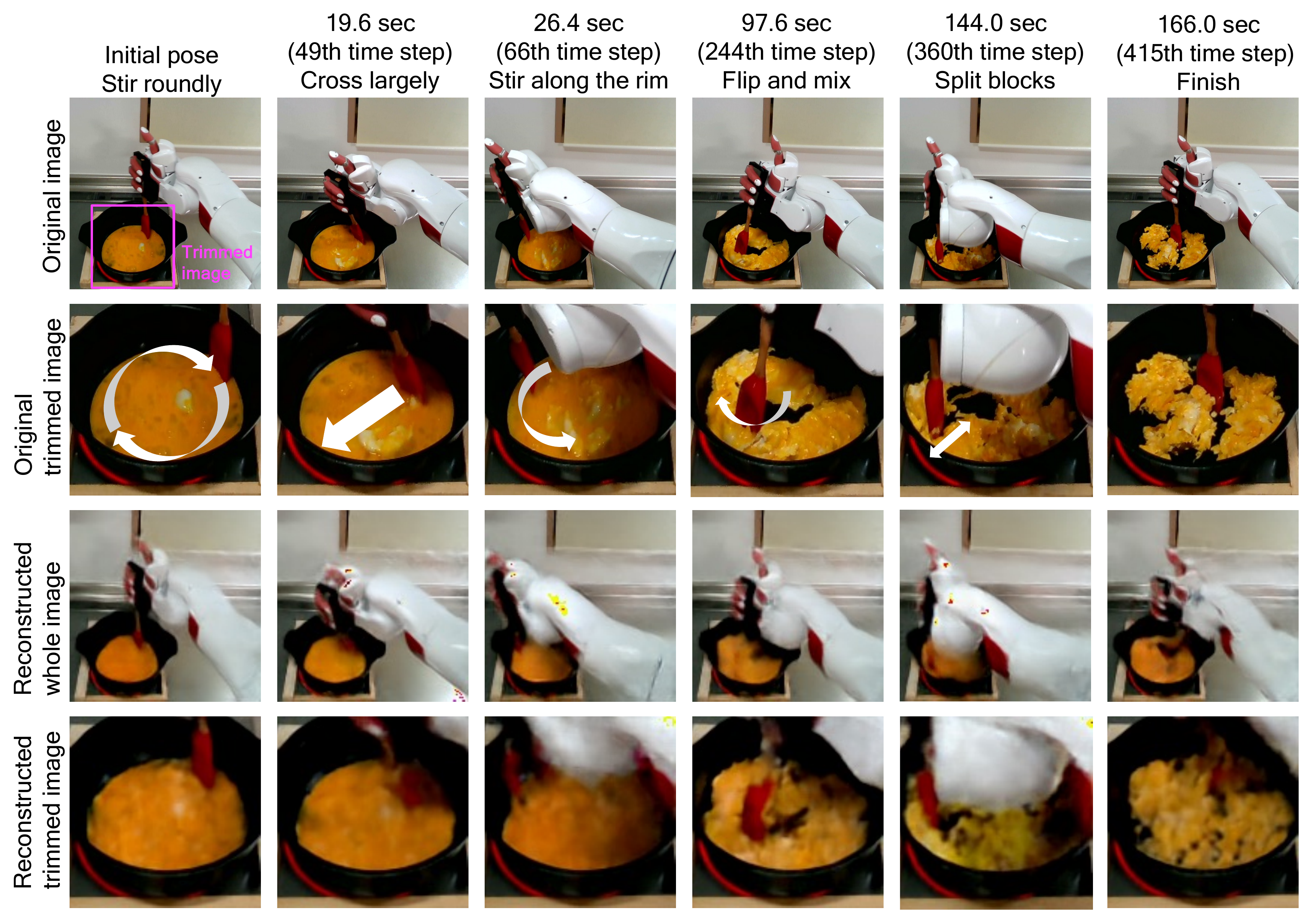}
\caption{
The robot generated motions with our model and succeeded in cooking using plain eggs, heating with 190${}^\circ$C. 
The original images and reconstructed images are shown. 
The purple squared areas are used as the trimmed images. 
The robot performed stirring, flipping, and splitting according to the situation.
}
\label{f:motion_generation}
\vspace*{-0.2cm}
\end{figure*}

Fig.~\ref{f:motion_generation} shows that the robot cooked plain eggs at a temperature of 190${}^\circ$C with the motion generated by our model.
First, as shown in the pictures, in the beginning, the robot stirred in a circle along the rim of the pot, expecting the robot to recognize the properties of the egg. 
Second, in steps 49 and 66, while the egg was still raw, the robot stirred largely all over the pan.
Subsequently, after the egg started being heated, the robot started flipping motion in the 244th step.
Next, when almost all parts had hardened in the 360th step, the robot cut the large egg block and split them.
Finally, in the 415th step, the arm returns to the initial pose and finished cooking.
These motions are not explicitly indicated, but the learning model learned them implicitly from demonstrations.

We also show the whole/trimmed images reconstructed by the CAE.
The whole images show and represent the arm postures, and the trimmed images show the status of the egg.
In particular, the trimmed images show which part of the egg is gathered, separated, or connected.
Thus, we assume that the CAE recognized the robot embodiment and the status of the egg.

\subsection{Cooking different dishes}
\begin{figure}[t]
\centering
\includegraphics[width=3.in]{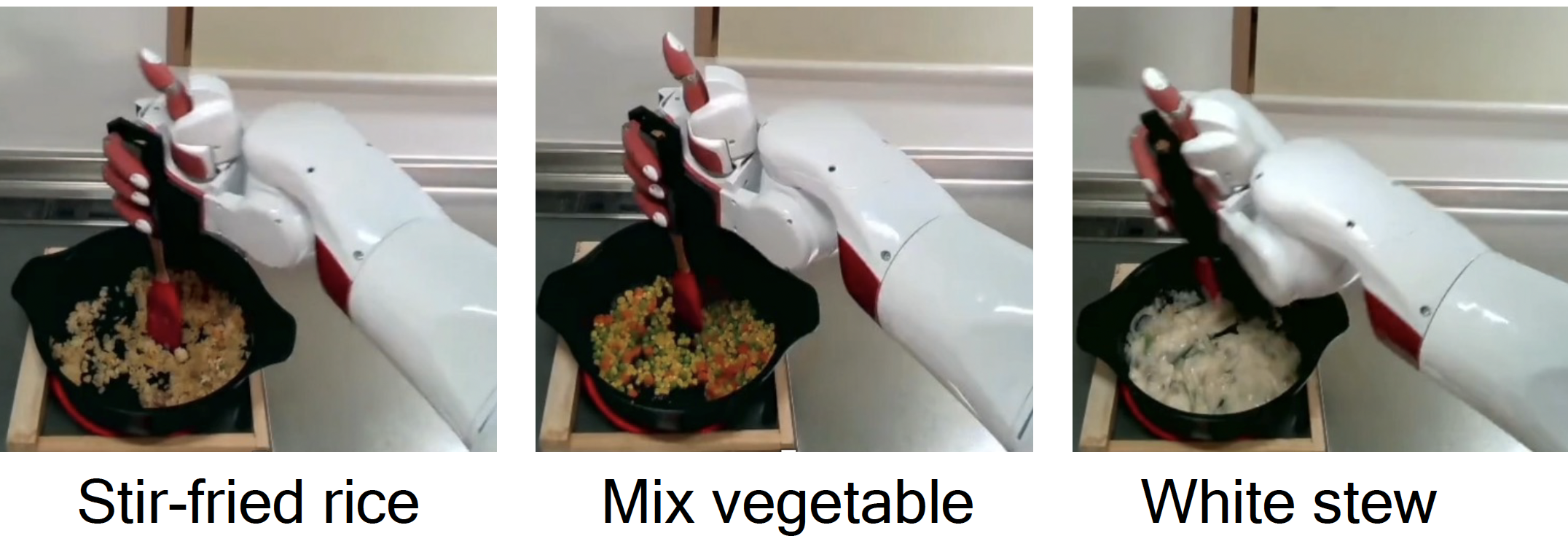}
\caption{Additional experiment with different dishes.
}
\label{f:different}
\vspace*{-0.2cm}
\end{figure}

Additionally, we tried cooking totally different dishes, as shown in Fig.~\ref{f:different}. 
The robot demonstrated stirring stir-fried rice, mixed vegetables, and white stew.
Although the states of the ingredients change differently from those of the egg, the robot showed movements approaching and separating the gathering parts.
The robot decided on the end timing of the motion, that is, when the stir-fried rice and vegetables become softer, and the white stew became smoother.
Therefore, our model potentially recognizes the transition of states of general targets.

\begin{figure*}[t]
\centering
\includegraphics[width=4.8in]{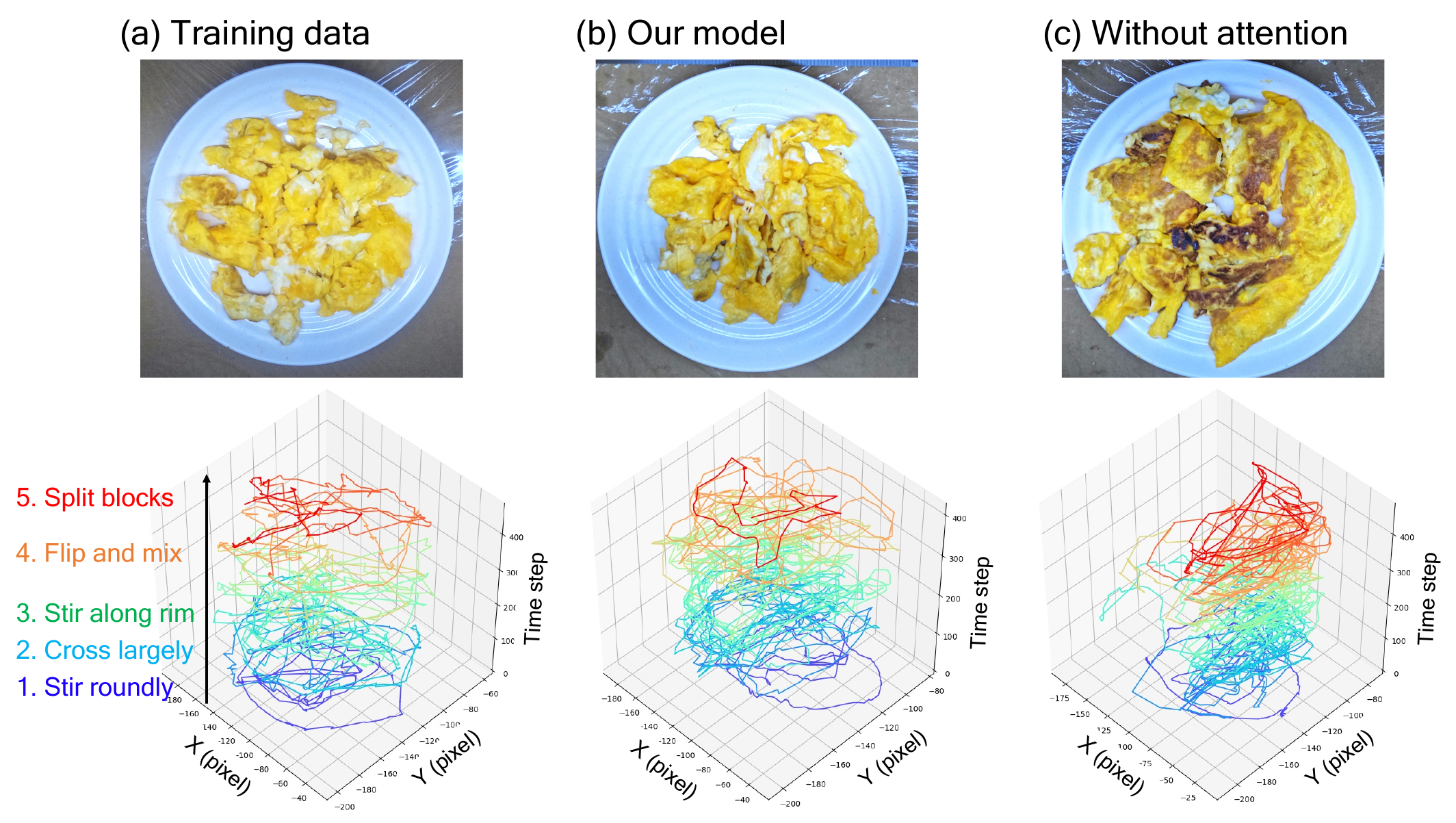}
\caption{
Arm trajectories during cooking with plain eggs at 190${}^\circ$C.
Stirring paths and final dishes are shown for training data, our model, and a model without attention.
Our model covered the pot broadly, and show a shift from wide stirring to focused, targeted motions over time, that is similar to the training data. 
While the no-attention model favored the right side to keep the pot visible.
}
\label{f:kinovea}
\vspace*{-0.2cm}
\end{figure*}

\begin{figure}[]
\centering
\includegraphics[width=3.4in]{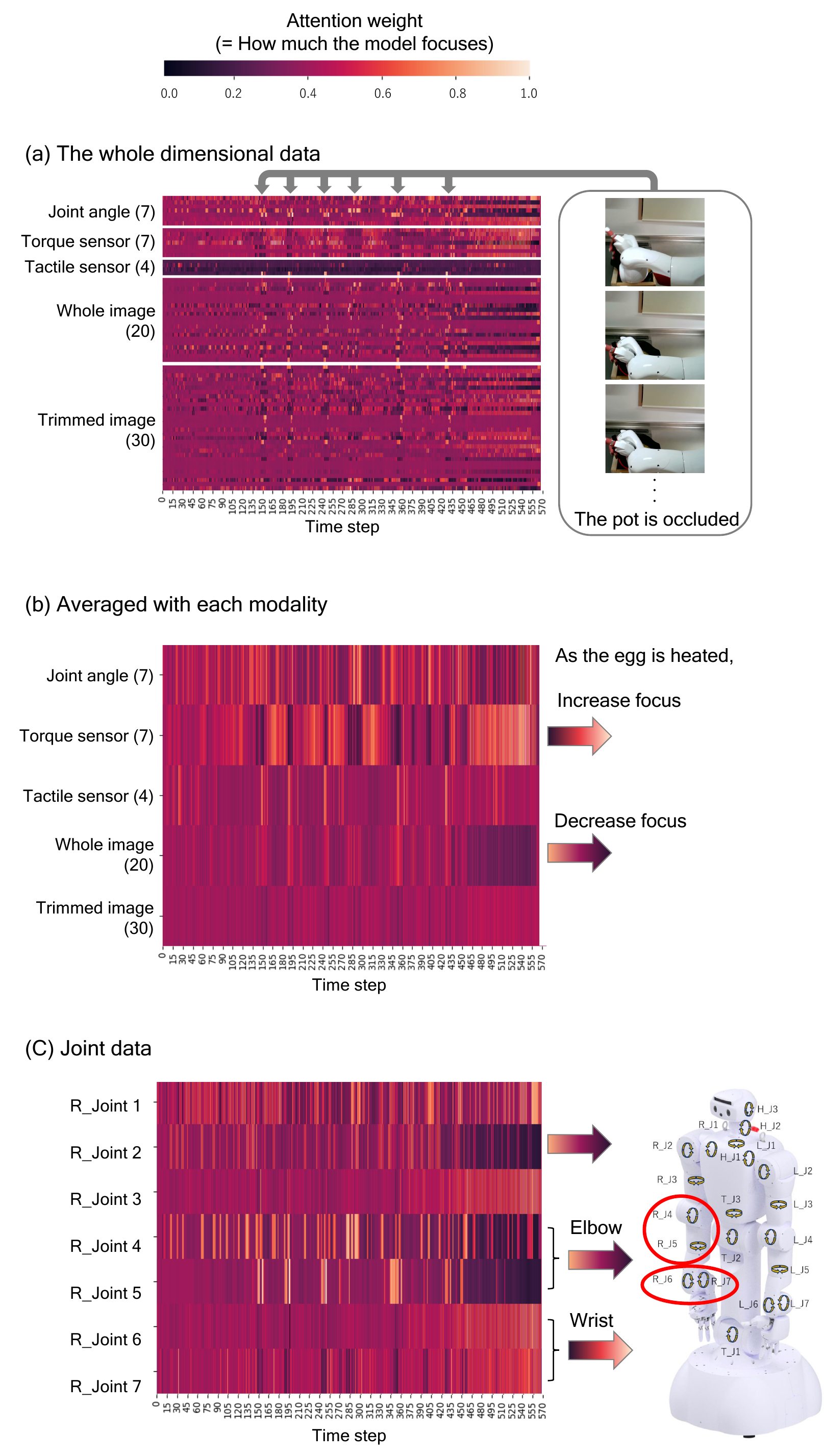}
\caption{
Attention maps during cooking with plain eggs at 180${}^\circ$C.
In (A), attention shifts across modalities occur with occlusion events, which are represented with vertical wave lines.
In (B), torque sensor gathers focus in the latter half, while whole image loses it.
In (C), Attention on joints 2, 4, and 5 decreases (used for wide stirring), while joints 6, and 7 increase (used for fine splitting).
}
\label{f:attention}
\end{figure}

\begin{figure}[]
\centering
\includegraphics[width=3.3in]{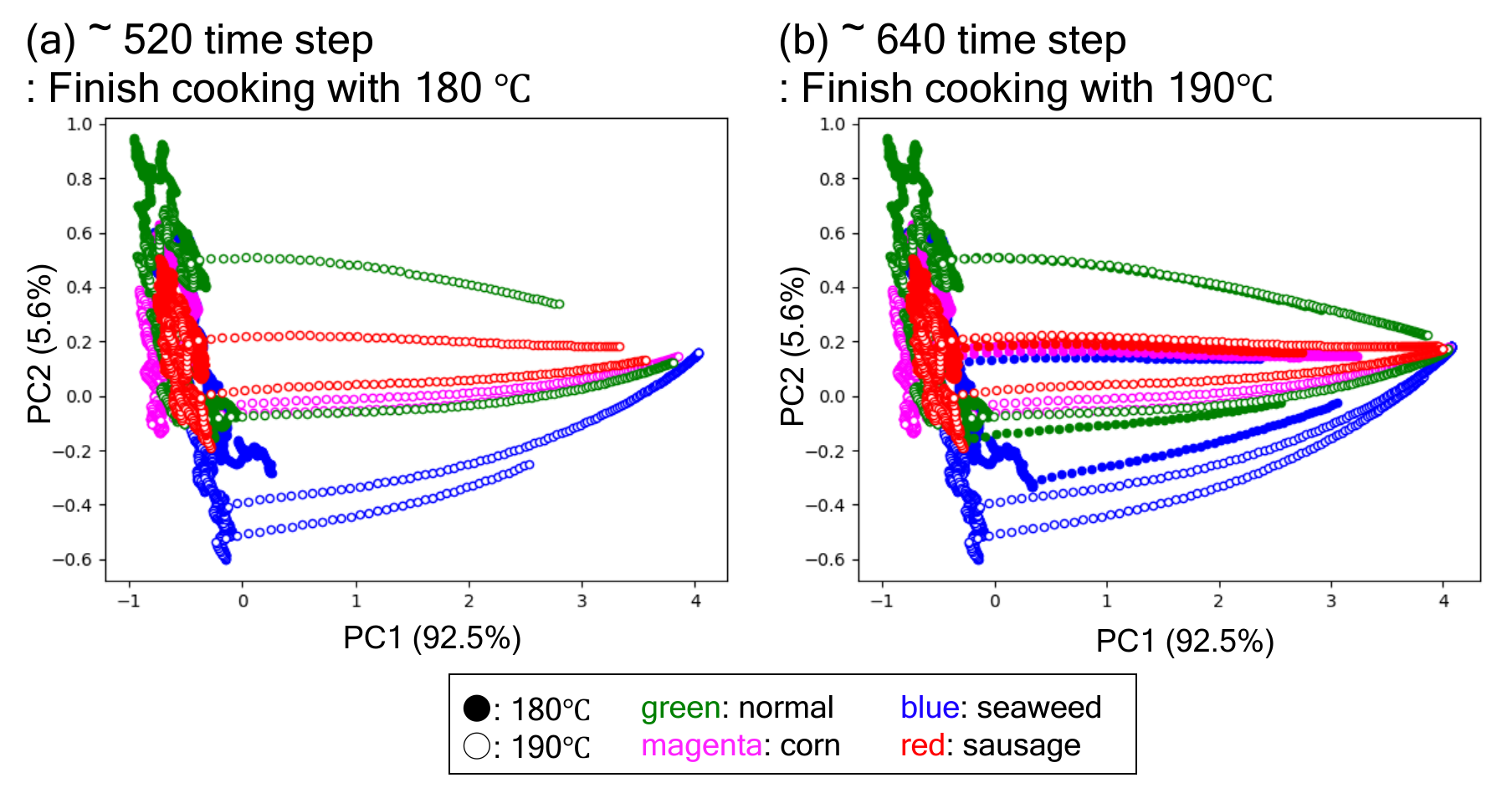}
\caption{
PCA of sequential Cs node values during cooking.
Two samples are shown for each heat power and ingredient.
Top: sequences up to 520 steps; Bottom: up to 640 steps.
PC1 captures the egg’s transition from raw to cooked, with values progressing along the PC1 axis.
The Cs node encodes stirring completion timing based on cooking status.
}
\label{f:PCA}
\end{figure}

\section{Discussion}
\subsection{Advantages of our model}
The proposed learning model offers three main advantages: real-time responsiveness, memory efficiency, and explainability for a variety of sensor modality.
First, our model provides real-time responsiveness, which is essential for dynamic tasks that require immediate feedback.
Many conventional methods such as the Action Chunking with Transformers (ACT) model~\cite{Fu2024, Zhao2023} generate sequences using time windows, which introduce latency.
In contrast, our predictive model outputs a single timestep prediction in each control cycle, enabling immediate adaptation to environmental changes.

Second, our model is memory efficient. 
It requires only 8.11GB for motion generation in this task, whereas ACT needs 29.9GB for the same setting. 
This makes our model suitable for edge computing in physical robots, which would be particularly beneficial for domestic assistive robots.

Finally, our model uses various sensory modality; not only vision as in many conventional research~\cite{Fu2024, Zhao2023, Chi2023, chi2024, Oct2023, Li2024} but also tactile and force; that enable handling intrinsic and extrinsic properties.
Moreover, our model offers modality explainability through the modality-specific attention mechanism, which has not been tackled before. 
This allows for a detailed understanding of how the robot interacts with objects whose state is changing, improving adaptability in dynamic tasks, and enhancing the robot's ability to make informed decisions in real-time.

\subsection{Analysis of arm trajectory comparing with ablation}
To confirm the effectiveness of our model, we conducted an ablation study.
We trained the same CAE and MTRNN architecture with the same parameters after removing the sensor modality attention mechanism from scratch.
Without the attention mechanism, we controlled the robot to cook plain eggs by heating at 190${}^\circ$C, over five trials, which is the condition under which all trials succeeded with our model.
As a result, in four trials, the robot pushed and hit the wall of the pot with too much force and stopped movement because it exceeded the limit of safety torque.
In one trial, the robot only stirred the right area in the pot, where the arm did not occlude the pot and the egg was visible.
Thus, without the attention mechanism, the robot cannot focus on torque and tactile sensor information, making it difficult to return to the correct motion just after the turner hits the pot wall.
Moreover, the model without an attention mechanism focuses on vision too much, and thus the robot only stirred the observable position.
We stopped the experiments for safety reasons because the hitting motions would damage the robot.

We analyzed the stirring trajectories, tracking the tips of the turner and compared the training data, our model, and the model without attention, as shown in Fig.~\ref{f:kinovea}.
They are all in the cases of the robot stirred plain egg with heating 190${}^\circ$C.
With our model shown in the middle figure of Fig.~\ref{f:kinovea}, the trajectory covers all the area in the pot as the training data on the left.
In addition, the trajectory shows that at first, the robot stirs roundly with large motions to approach the whole area.
Then gradually the motion was changed to sharp to flip or split, targeting the specific areas.
Finally the robot could cook scrambled egg as nicely as the training data.
On the other hand, the right figure, the model without attention mechanism, shows the robot only stirred the right area of the pot.
In the end, some large blocks were not stirred on the left side and burned.
Therefore, we can say that the modality attention mechanism is critical to adjust the movement depending on the objects' complex changing properties.

\subsection{Analysis of sensor modality attention}
We analyzed the attention mechanism to confirm the focused modalities each time.
In Fig.~\ref{f:attention}, we show attention maps, which are heat maps representing the attention weight.
The values in these figures are normalized and the lighter color indicates a highly focused modality, that the learning model recognizes as important and reliable.
They are all in the same case of the robot stirred plain egg with heating 180${}^\circ$C.

Fig.~\ref{f:attention} (a) is the attention map of the whole dimensional input data. 
We observe lines around the steps 148, 186, 242, 300, 347, and 425th, when occlusion occurred.
In other words, we see the transition of focused sensorimotor data.
The learning model recognizes the occlusion and changes the focus depending on the situation.

Fig.~\ref{f:attention} (b) shows the attention map averaged with each modality.
As can be seen, especially the heat color of torque sensor data becomes lighter and lighter, which means it attains focus in the latter half of the sequence.
We consider that because in the latter half, the egg becomes harder; thus, the learning model must focus on recognizing stiffness; which part is liquid, half-raw or hard.
In contrast, the attention towards the whole image data diminishes in the last part.
In the beginning, the robot has to move largely to stir the whole pot, but later, the movement becomes smaller and involves flipping or splitting specific areas, which mainly needs wrist transitions.
Thus, the whole image becomes less and less meaningful.

Fig.~\ref{f:attention} (c) shows the attention map that focuses only on motor angle data.
We see that joints 2, 4, and 5 decrease attention, which are on the shoulder and elbow, mainly used to stir in the whole pot. 
In contrast, joints 6, and 7 on the wrist increase attention, which are used to split the egg.
We can say that the attention mechanism focuses on important joints based on the status of the egg and the necessary action type.

\subsection{Analysis of egg status detection}
We conducted principal component analysis (PCA) on the sequential internal value of Cs ($u_{\rm {Cs}}[t]$) in the MTRNN while conducting the evaluation experiment. 
There are seven nodes in the Cs node, but with PCA, we can analyze the information represented in Cs with two dimensions.
Fig.~\ref{f:PCA} shows two samples for each heat power and ingredient; figure (a) shows the sequence until the 520th step, which is the timing only when heating with 190${}^\circ$C finishes cooking, and (b) shows until the 640th step, which is the timing of most trials with heating with 180${}^\circ$C also finishes cooking. 
In (a), with 190${}^\circ$C, all the flow has reached the plus side; however, with 180${}^\circ$C trials, all the flow has not moved from the minus side.
On the other hand, in (b), almost all the flow has reached the minus side.
Therefore, the transition from minus to plus in the PC1 axis explains the egg status, from raw to hard.
In conclusion, the Cs node represents the egg status and provides the complete stirring time, enabling the robot to complete the cooking task.

\section{Limitations and Future works} 
There are some areas for future work.
First, we aim to generalize this method to multiple sequential tasks. 
In this study, we focus only on a single manipulation task. 
However, many real-world tasks require combining multiple actions while adapting to dynamically changing object states. 
Recent studies using task planning frameworks~\cite{Yi2022, Wang2021} and foundation models~\cite{Oct2023, Li2024} have tackled long-horizon manipulation and could offer valuable insights.

Second, performing all experiments with a real robot in a physical environment presents practical limitations. 
To improve scalability and flexibility, we will utilize a simulation environment for safe and efficient data collection, and to establish a simulation-to-real transfer pipeline that ensures reliable deployment in real settings.

Third, future work includes more detail investigation for how the performance scales with additional sensor modalities or higher-resolution inputs.
Also we will evaluate the robustness of the model to sensor noise or failure in one or more modalities.

\section{Conclusion} 
\label{sec:conclusion}
We addressed the challenge of manipulating objects whose properties dynamically change.
We proposed a learning model that perceives both extrinsic and intrinsic object properties in real time, using multimodal sensory data that include vision, tactile, and torque information.
To achieve this, we introduced a predictive recurrent neural network with a modality attention mechanism that adaptively weighs sensory inputs based on their reliability and importance, enabling the robot to flexibly change focus according to the task context.
Through experiments involving a dynamic manipulation task, we demonstrated that the proposed system achieved robust and generalized performance even under untrained conditions.
The results highlight that selectively integrating multimodal information is critical for handling objects with continuously changing states, and that our approach can improve the real-time perception and adaptive control required in such tasks.
This work represents a step toward more autonomous and versatile robot systems capable of assisting in complex, dynamic, and uncertain real-world environments.

\bibliographystyle{IEEEtran}
\bibliography{IEEEabrv,references}
\end{document}